\documentclass[conference]{IEEEtran}
\IEEEoverridecommandlockouts
\usepackage{cite}
\usepackage{amsmath,amssymb,amsfonts}
\usepackage{algorithmic}
\usepackage[ruled,vlined]{algorithm2e}
\usepackage{graphicx}
\usepackage{textcomp}
\usepackage{xcolor}
\usepackage{comment}
\usepackage{hyperref}
\hypersetup{
    colorlinks=true,
    linkcolor=blue,
    filecolor=magenta,      
    urlcolor=cyan,
}
 
\def\BibTeX{{\rm B\kern-.05em{\sc i\kern-.025em b}\kern-.08em
    T\kern-.1667em\lower.7ex\hbox{E}\kern-.125emX}}
\begin{document}

\title{A Computationally Efficient Pipeline Approach to Full Page Offline Handwritten Text Recognition}

\author{\IEEEauthorblockN{Jonathan Chung}
\IEEEauthorblockA{
\textit{Amazon Inc.}\\
Vancouver, BC\\
jonchung@amazon.com}
\and
\IEEEauthorblockN{Thomas Delteil}
\IEEEauthorblockA{\textit{Amazon Inc.} \\
Vancouver, BC\\
tdelteil@amazon.com}
}

\maketitle

\begin{abstract}
Offline handwriting recognition with deep neural networks is usually limited to words or lines due to large computational costs.
In this paper, a less computationally expensive full page offline handwritten text recognition framework is introduced.
This framework includes a pipeline that locates handwritten text with an object detection neural network and recognises the text within the detected regions using features extracted with a multi-scale convolutional neural network (CNN) fed into a bidirectional long short term memory (LSTM) network.
This framework achieves comparable error rates to state of the art frameworks while using less memory and time. The results in this paper demonstrate the potential of this framework and future work can investigate production ready and deployable handwritten text recognisers.
\end{abstract}


\section{Introduction}
Due to the heterogeneous nature of handwritten text, it is much more difficult to automatically recognise compared to printed text (two orders of magnitude of difference in error rate, see Table 3 vs Table 5 in \cite{yousef2018accurate}). Recent successes in handwriting recognition can be attributed to developments in deep neural networks.
However, due to large computational costs, the systems are usually limited to recognising characters, words, and lines.
We propose a full page offline handwriting recognition framework that is less computationally expensive compared to existing frameworks.

\section{Related work}
\subsection{Text localisation}
Text localisation is an essential component of document layout analysis and accurate text localisation is crucial for handwriting recognition \cite{renton2018fully}.
Handcrafted features that utilise blob detection, clustering, edge detection, and histogram projections dominate in the traditional techniques.
More recently, data-driven techniques are becoming more prominent with the growth of neural networks.
Such techniques can be categorised based on the method in which the position of the text is defined. 
This includes lines \cite{gruning2018read}, bounding boxes \cite{moysset2015paragraph}, or areas containing ``text pixels" \cite{renton2018fully}.

In this paper, we predict bounding boxes around the text using deep learning techniques of object detection.
Given an image that contains multiple objects, object detection identifies bounding boxes that encompass the objects along with the confidence of the class of the object. 
In this work, the Single Shot MultiBox Detector (SSD) \cite{liu2016ssd} framework was applied to text localisation. 

\subsection{Text recognition}
\label{text_reg_review}
Handwritten digit recognition with the MNIST dataset was among the first work in deep learning \cite{lecun1990handwritten}.
However, the learning problem was limited to images of single digit characters.
Significant advances in handwritten text recognition were realised by the description of the multidimensional recurrent neural networks (MD-RNN) in Graves et al. \cite{graves2009offline} and the Connectionist Temporal Classification (CTC) loss.
A number of advances based on the MD-RNN were reported including using attribute embeddings \cite{toledo2017handwriting}, dropout \cite{pham2014dropout}, Tucker decomposition \cite{ding2017compact} etc.
Recent works conducted by Puigcerver \cite{puigcerver2017multidimensional} suggest that the multidimensional aspects of the MD-RNN can be replaced with feeding image-features (from a CNN) into a one-dimensional LSTM to significantly reduce the memory requirements of the systems.
The described methods are either limited to single words or single lines of handwritten text.
Bluche et al. \cite{bluche2016joint, bluche2017scan} described an end-to-end system that uses an MD-RNN along with an LSTM to encode multiple lines of text.
Although the described system shows promise to automatically recognise multiple lines, it may not be a practical solution as it requires a large amount of computational power \cite{puigcerver2017multidimensional}.
Winglinton et al. \cite{wigington2018start} utilised a region proposal network to find the starting positions of text lines and a line follower network was trained to trace the line of text.
This was followed by using a CNN-LSTM approach to recognise the characters.

\begin{figure*}
\centerline{\includegraphics[width=\textwidth]{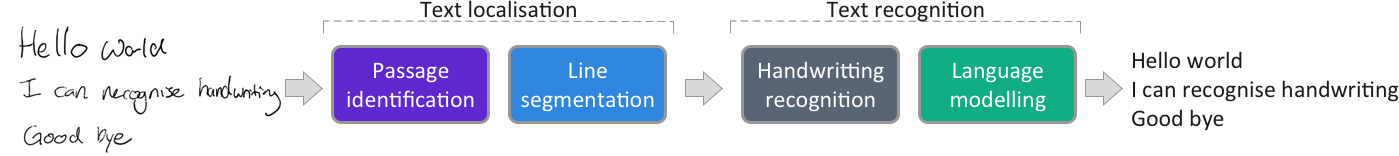}}
\caption{Overview of the system described.}
\label{fig1}
\end{figure*}

\subsection{Approach overview}
Previous works showed that the MD-RNN requires a large amount of computation power when it is used to recognise multiple lines of handwritten text.
A less computationally expensive alternative could be realised if multiple lines of handwriting recognition were not directly performed. 
To achieve this, our described framework is comprised of two major components: text localisation and recognition.
Text localisation identifies the positions of handwritten text given an image of the full page.
Once a passage of handwritten text was identified,  segmentation was conducted to locate each line of text.
Text recognition refers to converting an image of a line of handwritten text into a string with the corresponding characters and denoising the string with a language model.
By limiting handwriting recognition to single lines, the computational costs associated with this framework can be dramatically reduced compared to previous works that utilise the MD-RNN.

\section{Methods}

Rather than designing an end-to-end network, we took a modular approach consistent with described components in the literature.
This principle allows components of the framework to be easily replaced and tested with different ones.
An overview of our system is provided in Figure \ref{fig1}.

\subsection{Text localisation}
The purpose of text localisation is to identify bounding boxes of each line of text given an image containing both printed and handwritten text.
The text localisation procedure consists of two stages: passage identification and line segmentation.

\subsubsection{Passage identification}
\label{pas_iden}
The goal of passage identification is to predict the location of the handwritten passage (bounding boxes containing $x$, $y$ coordinates, width and height of the bounding box in percentages of the page size).
To simplify this step, we assume that there is one passage of printed text and one passage of handwritten text (using the IAM dataset \cite{marti2002iam} see Section \ref{eval} for more details).
This was achieved by extracting image features from a pre-trained truncated 34 layer residual network (ResNet34) \cite{he2016deep} trained on ImageNet.
In the ResNet34, the weights of the first convolutional layer were averaged into one channel to accommodate for greyscale images.
The features were then fed into three fully connected layers: two layers with 64 units and a relu activation and one layer with 4 units and a sigmoid activation.
The four units with sigmoid activation correspond to the $x$, $y$ coordinates, width, and height of the bounding box in percentages.
The network was trained to minimise the mean squared error.

\subsubsection{Line segmentation}
\label{line_seg}
Given an image containing only handwritten text, this component predicts bounding boxes surrounding each line of text.
We modelled this as an object detection problem to detect words followed by using a clustering algorithm to combine words into lines.
A two stage approach was taken because early experiments showed that the network was prone to missing objects when identifying handwritten text.
By detecting individual words, the chance of the network missing an entire line of words was less likely.

In our implementation, the SSD architecture \cite{liu2016ssd} was used to predict bounding boxes relative to anchor points and predict the probability that the bounding boxes are encompassing words.
The downsampler consists of two convolutional layers, batch normalisation layer, and a relu activation function.
The class and bounding box predictor consists of a single convolutional layer with 6 (4 positional + 2 for classes) output channels.
To adapt the SSD to our requirements, image features were extracted with a similar network described in Section \ref{pas_iden} (ResNet34).
Furthermore, the anchor boxes were adapted to resemble words (only squares and rectangles with widths $>$ height).
The SSD was trained to minimise the cross-entropy loss for the class (handwriting or not handwriting) and the L1 loss for the bounding box.
Non-maximum suppression was performed to filter out objects overlapping bounding boxes.

After the bounding boxes of words were detected, a greedy algorithm was then used to cluster the words into lines proposals based on the overlap in the y-direction (see Algorithm 1).

\begin{algorithm}
\SetAlgoLined
\textbf{Input}: \emph{words} a list of bounding boxes (x, y, w, h)\\
\textbf{Output}: \emph{lines} a list of bounding boxes (x, y, w, h)

Sort words based on the y-axis\\
\texttt{prev\_bb} = \emph{None}, \texttt{line\_bb} = []\\
\For{\texttt{bb} in \texttt{sorted\_words}}{
  \uIf{\texttt{prev\_bb} is not \emph{None}}{
    \eIf{\texttt{get\_y\_overlap(\texttt{prev\_bb}, \texttt{bb}) > 0.4}}{
        \emph{\#If the word in \texttt{bb} is within the same line}\\
        \texttt{line\_bb}.append(\texttt{bb})
    }{
        \emph{\#If the word in \texttt{bb} is in a different line}\\
        \texttt{output}.append(\texttt{get\_bb}(\texttt{line\_bb}))\\
        \texttt{line\_bb} = []
    }
   }
    \texttt{prev\_bb} = \texttt{bb}
 }
\texttt{output}.append(\texttt{get\_bb}(\texttt{line\_bb}))\\

\textbf{Helper functions}:\\
\texttt{get\_y\_overlap} - calculates the percentage overlap between two bounding boxes\\
\texttt{get\_bb} - gets the encompassing x, y, w, h coordinates given a list of bounding boxes\\

 \caption{Word to line clustering algorithm}
\end{algorithm}

The following heuristics were used to evaluate the line proposals:
\begin{itemize}
    \item Lines must have a minimum area
    \item Lines that exceed boundaries of the page are removed 
    \item Lines (excluding the last line) that are substantially shorter than the median width of the lines are removed 
    \item Lines that are much longer than the median height are split into 2 lines (accounts for double lines)
    \item Lines with starting positions that significantly deviate from other lines are removed
    \item Remove lines that greatly overlap with other lines
\end{itemize}

Lines that are not eliminated by the heuristics algorithm are used as the output of the text localisation stage. 

\subsection{Text recognition}
Text recognition takes images containing single lines of handwritten text and recognises the corresponding characters. 
Our approach includes handwriting recognition then denoising the output with a language model.

\subsubsection{Handwriting recognition}
\begin{figure}[t!]
\centerline{\includegraphics[width=0.5\textwidth]{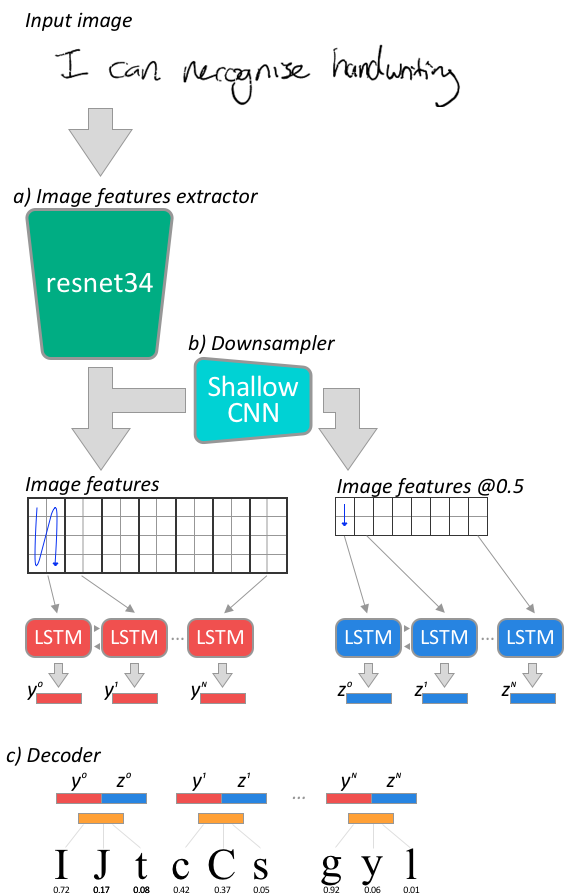}}
\caption{Handwriting recognition CNN-biLSTM.}
\label{fig2}
\centering
\end{figure}

Following a similar scheme to \cite{puigcerver2017multidimensional}, we implemented a CNN-biLSTM network. It makes use of a multi-scale CNN for image feature extraction, then the features are fed into a bidirectional LSTM. 
The network was trained to optimise the CTC loss (shown in Figure \ref{fig2}).
Intuitively, the CNN generates image features that are spatially aligned to the input image.
The image features are then sliced along the direction of the text to generate a fixed number of ``timesteps" and sequentially fed into an LSTM.

The CNN used to generate image features was identical to the residual network described in Section \ref{pas_iden} (ResNet34, Figure \ref{fig2}-a).
In order to account for varying sizes of the input image (e.g., lines that contain only one word compared to lines that contain seven words), multiple downsamples of the image features are provided (Figure \ref{fig2}-b, identical to the downsampler in the SSD used in Section \ref{line_seg}).
The image features and downsampled image features were each fed into separate biLSTMs.
The outputs of the biLSTMs were concatenated along the time dimension and decoded into a $N \times M$ array where $N$ is the maximum length of the sequence and $M$ is the number of unique characters (Figure \ref{fig2}-c).
This array is fed into the language model denoiser.
 
\subsubsection{Language model denoiser}
\label{denoise}
The $N\times M$ output of the CNN-biLSTM needs to be transformed into the output string.
As the output contains $N \times M$ probabilities corresponding to each character of the sequence, a naive solution (greedy solution) is to take the maximum probability (argmax) of each of the $N$ slices and collapse the characters using the CTC collapsing function.
Inspired by \cite{graves2009offline}, a beam search approach can alleviate such issues by combining multiple decoding paths to generate candidate strings. 
A language model can be included in the beam search decoding to weigh the proposals based on their likeliness.
The beam search approach required substantially more computational power as our early experiments, revealed that there was up to $11\times$ computational time increase compared to the greedy solution).

In this paper, a language denoiser network was developed.
Given a noisy input string, the network denoises the string in a sequence-to-sequence configuration.
A previous approach \cite{ghosh2017neural} encodes the noisy input at the character level and decode the clean output at the word level to ensure that the output is only composed of in-vocabulary words.
This has proved relatively effective however falls apart for out of vocabulary words like names and places.
To circumvent this issue, a character to character encoding / decoding scheme based on the Transformer architecture was used \cite{vaswani2017attention}.
The denoiser was trained on sentences from an external database of public domain novels \cite{publicdomainnovels}.
Characters are randomly inserted and deleted from the sentences (with a uniform distribution).
Also, characters are replaced with visually similar counterparts (e.g., `d' can be replaced with `c' and `l') in an attempt to model the real noisy distribution of the handwriting recognition model.
The generated noisy sentences are used to predict its original counterpart.
During inference, the output of the trained denoiser is fed into a beam search algorithm to generate candidate strings.
We make use of the following heuristics to rank them:
\begin{enumerate}
    \item Pick the candidate strings with the highest proportion of in-vocabulary words.
    \item Pick the candidate strings with the lowest Levenshtein distance.
    \item Pick the candidate string with the lowest perplexity score using an off-the-shelf pre-trained language model.
\end{enumerate}

\section{Experiment}
\subsection{Evaluation}
\label{eval}

The system was evaluated with the IAM dataset \cite{marti2002iam}.
The IAM dataset contains 1539 pages of scanned documents.
Each scanned document contains printed text and 657 writers were asked to write the contents of the printed text in the space provided.
The dataset was split into train and test data, where the test dataset includes validation 1, validation 2, and test data designated by the authors of the dataset.

We both evaluated the system qualitatively by visually evaluating the transcription of examples and quantitatively by computing the character error rate (CER).
Furthermore, we conducted memory and timing comparative analysis.

The CER was calculated with SCLITE \cite{fiscus_sclite_1998} and the effects of the following components were evaluated: 1) no line heuristics, 2) no language model (argmax algorithm), 3) with beam search \cite{graves2009offline}, and 4) with the denoiser described in Section \ref{denoise}.
The predicted and actual text were aligned and the average CER was calculated line-by-line.
Our method was compared to similar methods presented in
\cite{wigington2018start,bluche2017scan,bluche2016joint}.

\subsection{Training details}
The networks were developed with Apache's MXNet deep learning framework \cite{chen2015mxnet}.
The networks for each component (passage identification, line segmentation, handwriting recognition, and language model denoising) were trained separately and the Adam optimiser was used for all the networks \cite{kingma2014adam}.
Data augmentation including random translation, shearing, and occlusions were performed. However, many typical data augmentation are not applicable to this application (e.g., flipping and random cropping).
In the word/line object recognition component, to circumvent this issue, lines or words were randomly blanked out.
Details of the implementation can be seen here (\url{https://github.com/awslabs/handwritten-text-recognition-for-apache-mxnet}).

\section{Results}

\begin{figure*}
\centerline{\includegraphics[width=\textwidth]{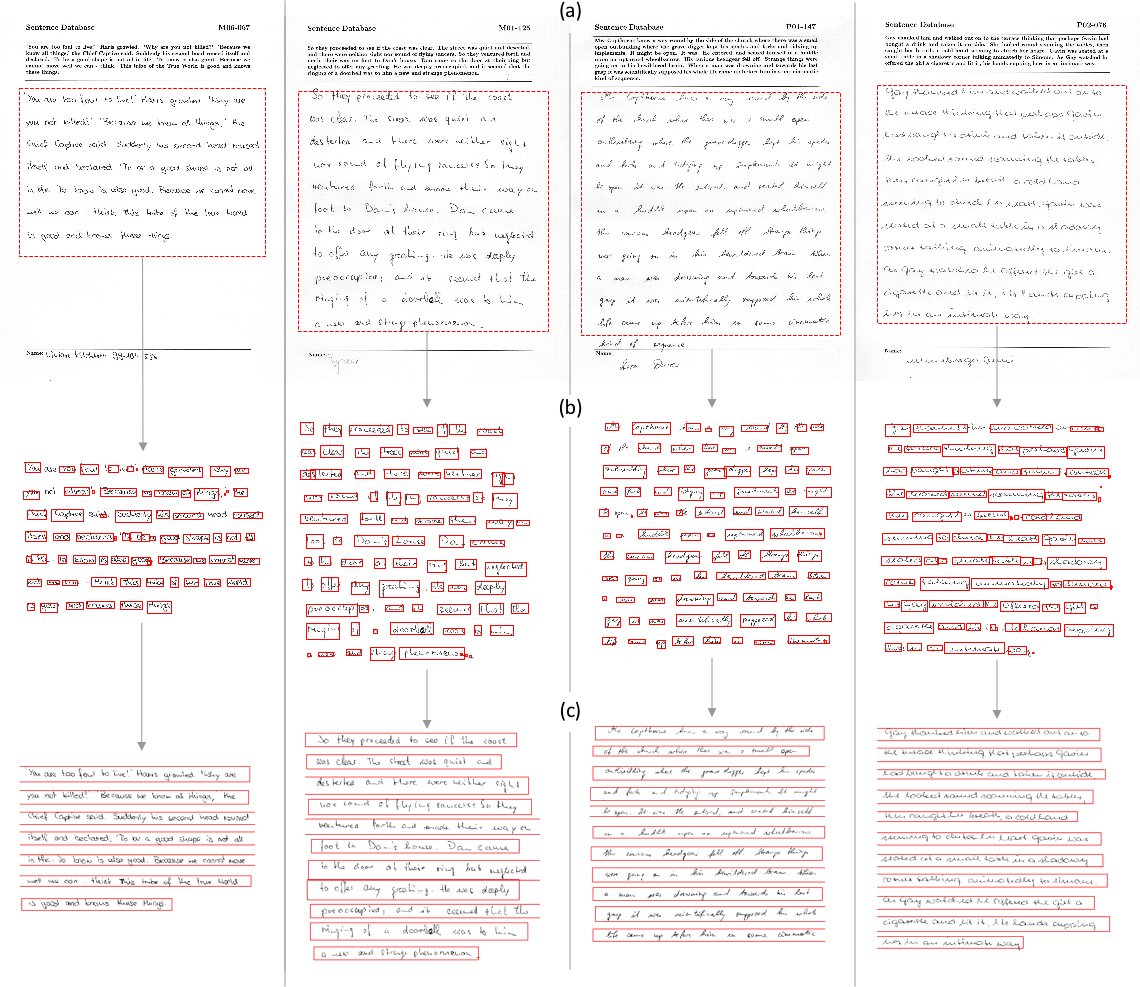}}
\caption{Qualitative results: full page to line images. a) paragraph segmentation, b) word segmentation, c) word to line conversion. The pipeline of the images goes from top to bottom within each column.}
\label{fig3}
\end{figure*}

Figure \ref{fig3} shows actual results of paragraph segmentation and word to line segmentation.
We can observe that the paragraph segmentation algorithm mostly predicts the bounding boxes of the handwriting component successfully, however, the third column presents a failure case where the last line is not encompassed by the predicted bounding box.
Given an image containing only handwritten text, the word detection algorithm can detect tight bounding boxes for each word.
However, as mentioned in Section \ref{line_seg}, there are several short words (typically with words $<$ 3 characters) that are not detected.
Despite the missing words, we can observe in Figure \ref{fig3}-c that all the lines were successfully detected. 

\begin{figure*}
\centerline{\includegraphics[width=\textwidth]{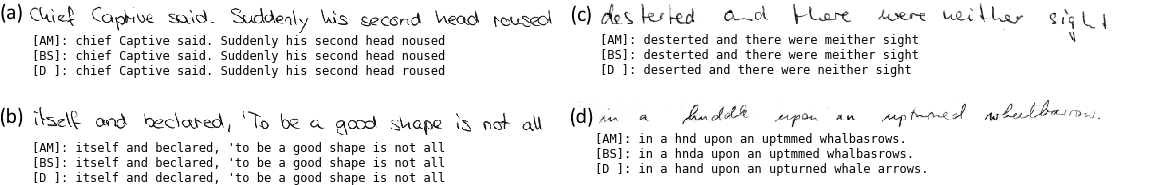}}
\caption{Handwriting recognition and language modelling. Four line images are displayed and under each line image contains the predicted string ([AM]: greedy argmax algorithm with no language modelling, [BS]: beam search \cite{graves2009offline}, [D]: our denoiser Section \ref{denoise})}
\label{fig4}
\end{figure*}

Figure \ref{fig4} presents the selected examples to show differences between the language model component of the described system.
First, we can observe that the greedy algorithm ([AM]) performs reasonably well and the beam search ([BS]) algorithm does not dramatically improve the results.
In a), we can see that the word ``noused" was converted into ``roused", which may be based on the preceding word ``head" and the visual similarity of `n' and `r'.
In b), the handwriting looks like ``beclared" but the denoiser replaced `b' with `d' based on the learnt language model.
In c), the `t' in ``desterted" was deleted also based on language modelling.
In d), none of the algorithms were successful to correct the sentences, and the denoiser worsened the CER.

The CER presented in Table \ref{tab1} suggests that line heuristics dramatically improved handwriting recognition.
Qualitatively evaluating the results suggest that the line heuristics algorithm eliminated incorrectly identified lines that caused large disparities when aligning the predicted and correct text.
The denoiser achieved a 1.4 CER decrease compared to the greedy argmax algorithm and beam search algorithm.
When compared to previous works on recognising cropped images (i.e., feeding a \emph{cropped} image containing only the handwritten portion compared to the \emph{full} page with printed and handwritten text, as indicated by \textbf{Seg.} in Table \ref{tab1}), our method outperforms Bluche \cite{bluche2017scan}. However, methods described in Bluche \cite{bluche2016joint} and Wigington \cite{wigington2018start} had lower CER compared to our method.

\begin{table}[htbp]
\caption{CER results}
\begin{center}
\begin{tabular}{l|c | c}
\textbf{Method} & \textbf{Seg.} & \textbf{CER}\\
\hline
Greedy argmax without line heuristics (ours) & Full & 28.3\\
Greedy argmax algorithm (ours) & Full & 9.90\\
Beam search algorithm (ours) & Full & 9.90\\
Denoiser (ours) & Full & 8.5\\
\\
Bluche \cite{bluche2016joint} & Cropped & 7.9\\
Bluche \cite{bluche2017scan} & Cropped & 16.2\\
Wigington \cite{wigington2018start} & Cropped & 6.4\\
\end{tabular}
\label{tab1}
\end{center}
\end{table}

\begin{table}[htbp]
\caption{Memory and timing requirements}
\begin{center}
\begin{tabular}{l|c | c}
\textbf{Method} & \textbf{Time taken (sec)} & \textbf{Memory (gb)}\\
\hline
Ours & 0.389 $\pm$ 0.0888 & 2.6  \\
Bluche \cite{bluche2016joint} & 21.2 $\pm$ 5.6 & N/a\\
Bluche \cite{bluche2017scan} & 0.62 $\pm$ 0.14 & N/a\\
Wigington \cite{wigington2018start} & 0.546 $\pm$ 0.447 & 9.7\\
\end{tabular}
\label{tab2}
\end{center}
\end{table}

Table \ref{tab2} presents the memory and timing requirements for our memory compared to existing methods.
When comparing the mean time taken to run an image, our method requires approximately $1.5\times$ less time compared to \cite{wigington2018start} and \cite{bluche2017scan}.
Our method also utilises substantially less memory (approximately $3.5\times$ less memory) compared to \cite{wigington2018start} (unfortunately, the memory requirements for \cite{bluche2016joint} and \cite{bluche2017scan} could not be attained).
Since our memory usage is substantially smaller, it is possible to run multiple images at the same time; effectively reducing the time required by a third.

\section{Conclusion}

In this paper, we presented a full page offline handwritten text recognition framework.
This framework consists of a pipeline where the handwritten text is localised (text localisation) followed by converting images of words into strings (text recognition).
Our method achieved a CER of 8.50.
The main advantage of the framework introduced is the reduced computational costs compared to existing methods.
For a tradeoff of CER$\approx$2 comparing to \cite{wigington2018start}, the throughput could be effectively $4.2\times$ when using a similar amount of memory.

In conclusion, the framework that we presented is a computationally cheap alternative to performing full page offline handwritten text recognition.
The results in this paper demonstrate the potential of this framework and future work can investigate different components of the pipeline for improved results. 

\section*{Acknowledgement}
Thank you Simon Corston-Oliver, Vishaal Kapoor, Sergey Sokolov, Soji Adeshina, Martin Klissarov, and Thom Lane for their helpful feedback for this project.

\bibliographystyle{IEEEtran}
\bibliography{references}

\end{document}